\documentclass{article}

\usepackage[preprint]{nips_2018}

\usepackage[utf8]{inputenc} 
\usepackage[T1]{fontenc}    
\usepackage{hyperref}       
\usepackage{url}            
\usepackage{booktabs}       
\usepackage{amsfonts}       
\usepackage{nicefrac}       
\usepackage{microtype}      
 
\usepackage{graphicx}
\usepackage{subcaption} 

\usepackage{amsmath}
\usepackage{amssymb}
\usepackage{mathrsfs}
\usepackage{yfonts}
\usepackage{bm}
\usepackage{multirow}

\usepackage{algorithm}
\usepackage{algpseudocode}

\usepackage[T1]{fontenc}
\usepackage[libertine,cmintegrals,cmbraces,vvarbb]{newtxmath}
\usepackage[scaled=0.95]{inconsolata}
\usepackage{fix-cm}

\include{Images}
\usepackage{color}
\usepackage{soul}
\usepackage{enumitem}
\usepackage[us]{datetime}

\usepackage[table]{xcolor} 

\newcommand{\beq} {\begin{equation}}
\newcommand{\eeq} {\end{equation}}
\newcommand{\beqs} {\begin{equation*}}
\newcommand{\eeqs} {\end{equation*}}

\newcommand{\sP}{\ensuremath{\mathcal{P}}}

\newcommand{\sX}{\ensuremath{\mathcal{X}}}
\newcommand{\sY}{\ensuremath{\mathcal{Y}}}

\title{Domain2Vec: Deep Domain Generalization}

\author{
  Aniket Anand Deshmukh, Ankit Bansal, Akash Rastogi\thanks{All three authors contributed equally.} \\
  Department of EECS\\
  University of Michigan\\
  Ann Arbor, MI 48105 \\
  \texttt{anikede, arbansal, arastog @umich.edu} \\
}

\begin{document}

\maketitle

\begin{abstract}
We address the problem of domain generalization where a decision function is learned from the data of several related domains, and the goal is to apply it on an unseen domain successfully. It is assumed that there is plenty of labeled data available in source domains (also called as training domain), but no labeled data is available for the unseen domain (also called a target domain or test domain). We propose a novel neural network architecture, Domain2Vec (D2V) that learns domain-specific embedding and then uses this embedding to generalize the learning across related domains. Proposed algorithm, D2V extends the idea of distribution regression \cite{poczos2012support, zaheer2017deep} and kernelized domain generalization \cite{deshmukhMDG2017} to the neural networks setting. We propose a neural network architecture to learn domain-specific embedding and then use this embedding along with the data point specific features to label it. We show the effectiveness of the architecture by accurately estimating domain to domain similarity. We evaluate our algorithm against standard domain generalization datasets for image classification and outperform other state of the art algorithms. 

\end{abstract}

\section{Introduction}
There has been significant interest in the last decade on the idea of generalization of learning, on whether it is possible to use similarities in underlying spaces and distributions \cite{deshmukhMDG2017}, \cite{muandet2013domain}, \cite{blanchard2011}, \cite{gretton2009dataset} or tasks \cite{caruana1998multitask} to build general learning algorithms to work with new settings. When working with datasets which have similarities in the underlying data distribution, it is possible to use the variation in the distributions to generalize learning tasks \cite{christmann2010universal}. 

In our scenario, we are given \(N\) datasets with each dataset drawn from different probability distributions such that each point in each dataset has a label or class associated with it. The goal is to learn a classifier such that, given a new dataset (with no training data/labels) drawn from a different but similar probability distribution, it is possible to provide labels to its points. The goal is to predict these labels corresponding to samples drawn from the marginal test distribution for the test dataset.  
 
Classification datasets can be associated with a common structure which allows us to generalize behavior. Consider the example of image classification. We have multiple images of same objects from different cameras, and we train our image classification model for these cameras (e.g., Apple's iPhone, Sony's DSLR, and Google's Pixel). During the test time, the goal is to classify images from Samsung's Note. Different cameras have different optical structure and images from those cameras may look slightly different or may have different optical properties. What makes this more difficult is that there are no labeled images from Samsung's Note and the classification model should be able to adapt without any labeled data. Instead of cameras, one could also think of different datasets VOC2007, LabelMe, Caltech-101, and  SUN09 (VLCS).

Our main contributions in this paper are: i) proposing the architecture for domain generalization and domain adaptation, ii) showing the effectiveness of the architecture by learning domain-specific embedding and thereby domain to domain similarity and iii) experimental evaluation of the architecture in different scenarios.

\section{Formal Problem Statement}
\label{sec:FormalProblemStatement}
Let  $\mathcal{X} $ be the feature space and $\mathcal{Y} $ be the label space. Let $\mathcal{P}_{ \mathcal{X} \times \mathcal{Y}} $ be the set of probability distributions on $\mathcal{X} \times \mathcal{Y}$ and $\mathcal{P}_{ \mathcal{X}}$ the set of probability distributions on  $\mathcal{X}$. Furthermore, it is assumed that there exists a probability measure $\mu $ defined over $\mathcal{P}_{ \mathcal{X} \times \mathcal{Y}} $. We are given sample points from N distributions: $ S_i = ((X_{ij}, Y_{ij}))_{1 \leq j \leq n}$ such that \((X_{ij},Y_{ij}) \sim P_{XY}^i\) and \(P_{XY}^i \sim \mu\) $, \forall i = \{1,...,N \}  $. For a new test dataset $S^T = (X_{j}^T)_{1 \leq j \leq n_T}$ and \(P_{X}^T \sim \mu_X\), the goal is to accurately predict \((Y_j^T)_{ 1\leq j\leq n_T }\). Let \(\tilde{\mathcal{X}} =  \sX \times \sP_{\sX}\) be the extended feature space. In the domain generalization setting, a decision function is a function $f:  \sX  \times \mathcal{P}_{ \mathcal{X}} \rightarrow \sY $ that predicts $\hat{Y_j} = f(X_j,\hat{P_X}) $, where $\hat{P_X}$ is the associated empirical distribution. We aim to learn a decision function \(f\) that minimizes the expected loss  \(\mathbb{E}[\ell(f(\tilde{X}),Y)]\), where \(\ell:\sY \times \sY \rightarrow \mathbb{R}\) is a loss function. 

With the assumption that the marginal distribution $P_{X} $ of domain represents that particular domain, we propose a general neural network architecture to learn some statistical estimates (eg. moments of distribution) from the data of that domain. These estimates (domain specific embeddings) are then combined with actual data from the domain to learn a final classifier. This procedure is exactly same as learning a decision function $f:   \sX \times \mathcal{P}_{ \mathcal{X}} \rightarrow \sY $ over extended feature space \(\sX \times \sP_{\sX}\). We aim to learn a decision function \(f\) to minimize the expected loss  \(\mathbb{E}
[\ell(f(\tilde{X}),Y)]\). We define empirical training error, empirical test error and generalization error as,
\begin{itemize}
\item Empirical training error: 
\[ \hat{\varepsilon}(f) = \frac{1}{N} \sum_{i = 1}^{N} \frac{1}{n_i}  \sum_{i = 1}^{n_i} \ell (f(\hat{P_X^i},X_{ij}),Y_{ij}). \]
\item Empirical test error:
\[ \hat{\varepsilon}(f)^T = \frac{1}{n_T} \sum_{i = 1}^{n_T} \ell (f(\hat{P_X^T},X_i^T),Y_i^T). \]
\item Generalization error: 
\[\varepsilon (f) = E_{P_{XY}^T \sim \mu}  E_{(X^T,Y^T) \sim P_{XY}^T} \ell (f( P_X^T ,X^T),Y^T) . \]
\end{itemize}

 In the next section, we describe the proposed architecture to solve the problem of domain generalization. 
\section{Proposed Architecture}
The proposed architecture is motivated by the idea that one needs to find a classifier or a decision boundary from  $\tilde{\sX} = \sX \times \sP_{\sX}$ to  $\sY $. In a typical neural network for supervised learning, we just find a classifier from  $ \sX $ to $\sY $. The idea here is then to use a neural network to learn some statistic $ T_X $ or the empirical characteristic function of the marginal probability distribution $\sP_{\sX}$ and concatenate the learned vector $ T_X $ with a sample $X $ from  $ \sX $. Let $ \sX \in \mathbb{R}^d $ be the $d$ dimensional feature space and for the $i^{th} $ distribution let $S_x^i = \{X_{i1},...,X_{in} \} \in \mathbb{R}^{n \times d}$.  The typical neural network just operates on the dimension $d$ and transforms the original features $ S_x^i$ to some $ D$ dimensional features  $\Phi_x^i$. But this doesn't learn any statistic from $n$ samples. Our proposed idea is to operate on dimension $n$ and get the domain-specific statistic $ T_X^i $ of the data and use it along with the features $ S_x^i$. The idea is that for data point $X_{ij} $ to have same label as $X_{pq} $, not only does $X_{ij} $ need to be close to $X_{pq} $ but so should $ T_X^i $ to $ T_X^p $.

\begin{figure}[H]
  \centering
    \includegraphics[width=0.8\textwidth]{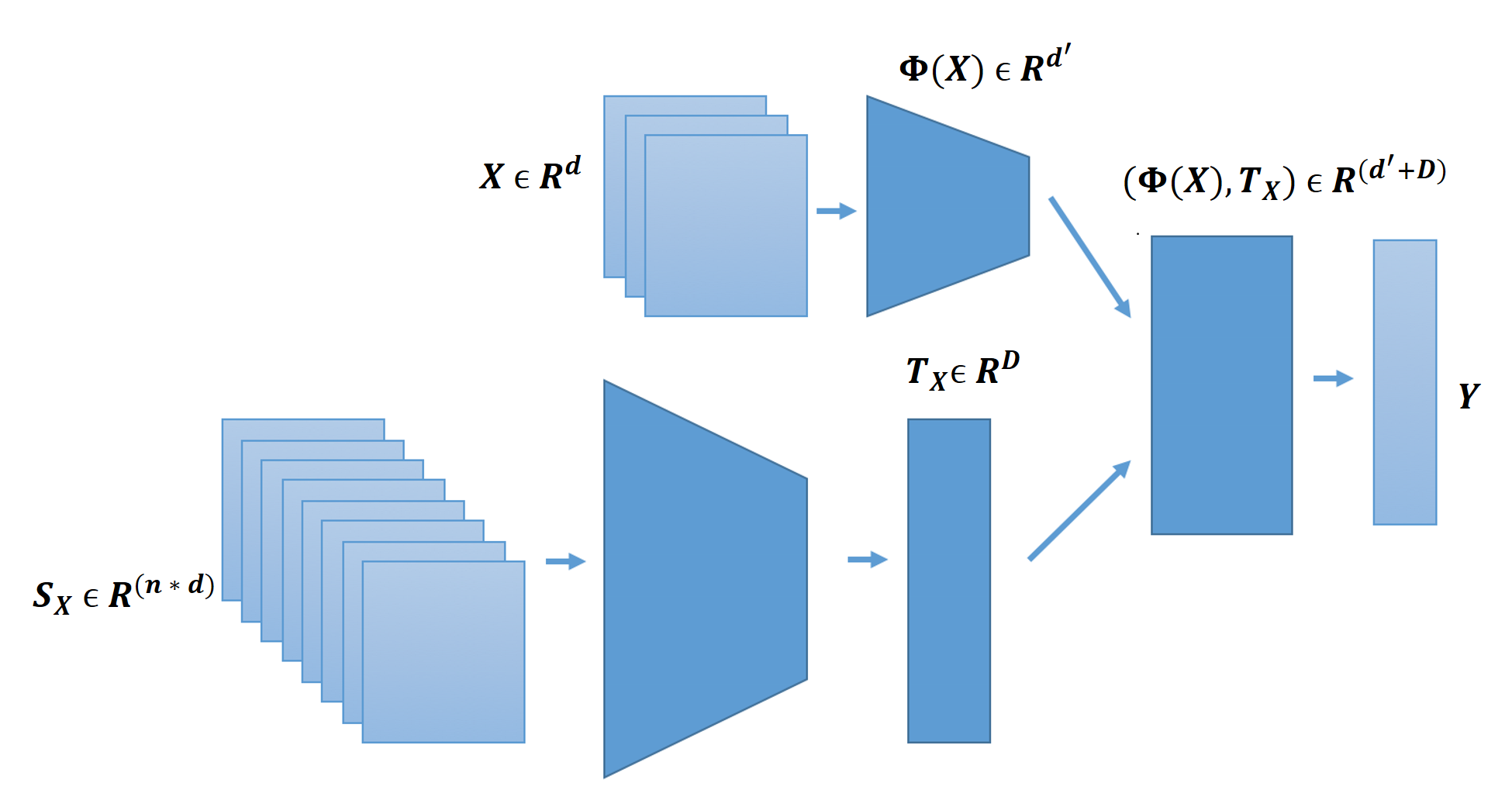}
     \caption{Domain generalization architecture by learning task specific embedding}
     \label{fig:proposed_architecture}
\end{figure}

\subsection{Related Work}
\subsubsection{Distribution Regression}

Our proposed method is closely related to distribution regression. The goal of distribution regression is to label a set of points (cloud of points) assuming that all points belonging to one set come from one distribution. Some of the applications of distribution regression are to estimate statistic of the population and to draw some conclusions based on the entire data (rather than seeing each point in the set individually). 
\cite{poczos2012support} and \cite{szabo2016learning} define the kernel on probability distribution and use kernel methods for classification or regression. Similarly \cite{bachoc2017gaussian} uses Gaussian process for distribution regression. \cite{edwards2016towards} uses variational encoders to estimate statistics of the data which can then be used for classification or regression.

\cite{zaheer2017deep} and \cite{law2018bayesian} learn the function from a set of points (cloud of points from features space) to the label space. This function is called permutation invariant because any estimated statistic of the set does not depend on the ordering of elements in the set. They also propose permutation equivariant method which is related to our proposed architecture but is very restricted. Our architecture \ref{fig:proposed_architecture} is very general and can be used for any domain generalization and multi-task learning application. Part of our proposed network is motivated by distribution regression, which we will use to get domain specific information.  

\subsubsection{Domain Generalization}
One of the methods to solve the transfer learning problem when there is no labeled data available for the test domain is described in \cite{deshmukhMDG2017} and \cite{blanchard2011}.  The approach is a distribution-free and kernel-based. We shall call this approach as kernelized domain generalization (KDG). It involves identifying an appropriate reproducing kernel Hilbert space (RKHS) and optimizing a regularized empirical risk over the space. The approach results in support vector machine defined over an extended feature space. This approach gives the generalization error bound in terms of number of datasets, number of examples per dataset and number of classes. Though the method is theoretically proven to be powerful, it is limited by the SVM. One needs to define the kernel that's useful for the dataset in hand and it isn't as powerful as neural networks when working with images and texts. 
 
Multi-task Auto Encoder (MTAE) learns domain invariant features using autoencoder setting. It has a single encoder and $D$ decoders, where D are the number of domains in the training data. At the end of autoencoder training, it learns features that are robust to variations across domains. The learned features are then used as inputs to a classifier. The classifier is trained on all the data from domains available during training \cite{ghifary2015domain}. Scatter Component Analysis (SCA) proposes a geometrical measure called scatter. Scatter helps to find a representation that minimizes the mismatch between domains and maximizes the separability of the data. Optimization in SCA can be reduced to generalized eigenvalue problem \cite{ghifary2017scatter}. Classification and contrastive semantic alignment (CCSA) learns the domain invariant embedding by minimizing the sum of classification loss, confusion alignment loss, and semantic alignment loss. Confusion alignment loss is a distance between distribution and a popular choice for such a loss is maximum mean discrepancy. Semantic alignment loss makes sure that samples from different domains and with different labels are mapped as far apart as possible in the embedding space \cite{motiian2017unified}. 
\cite{li2017deeper} this paper is similar to Undo-Bias \cite{khosla2012undoing}, where domain agnostic model (which benefits all domains) and domain-specific model is learned during training. When one observes a new domain, only domain agnostic model is used. Meta-learning domain generalization approach (MLDG) provides a model agnostic training procedure that improves the domain generality. It minimizes the training loss in such a way that virtual testing loss (which generalizes across different domains) is also minimized during the training phase \cite{li2017learning}. 
The proposed method D2V has several advantages over these existing methods. First, the model is easy to understand, implement and motivated by distribution or dataset similarity. Domain2vec yields domain specific embedding which can be used to represent a particular dataset and quantify the similarity between different datasets.

 \section{Experimental Results}
{
D2V can be viewed as comprising of two main components: 1) the \textit{main} model and 2) the \textit{task} model. In a rather unconventional setting, we operate on the batch size dimension in the tensors of the task network. The main network is fed a \textit{smaller} batch of data belonging to one task while the task network is fed with a \textit{larger} batch of data for the same task (typically as large as the entire task). The idea behind this practice is that D2V network learns task/domain specific information and hence 'sees' close to entire task at every iteration from the task network while also learning inter-domain nuances from the main network. Together, the two components of D2V enable it to learn robust embeddings across the domains, thereby improving the classification accuracy. For experiments in this paper, we use single hidden layer in both main network and task network. We get best hyper-parameters (learning rate, number of hidden units in the main network and task network, wight decay) using random search.  
}

 \begin{figure}[htb!]
  \centering
    \includegraphics[width=0.85\textwidth]{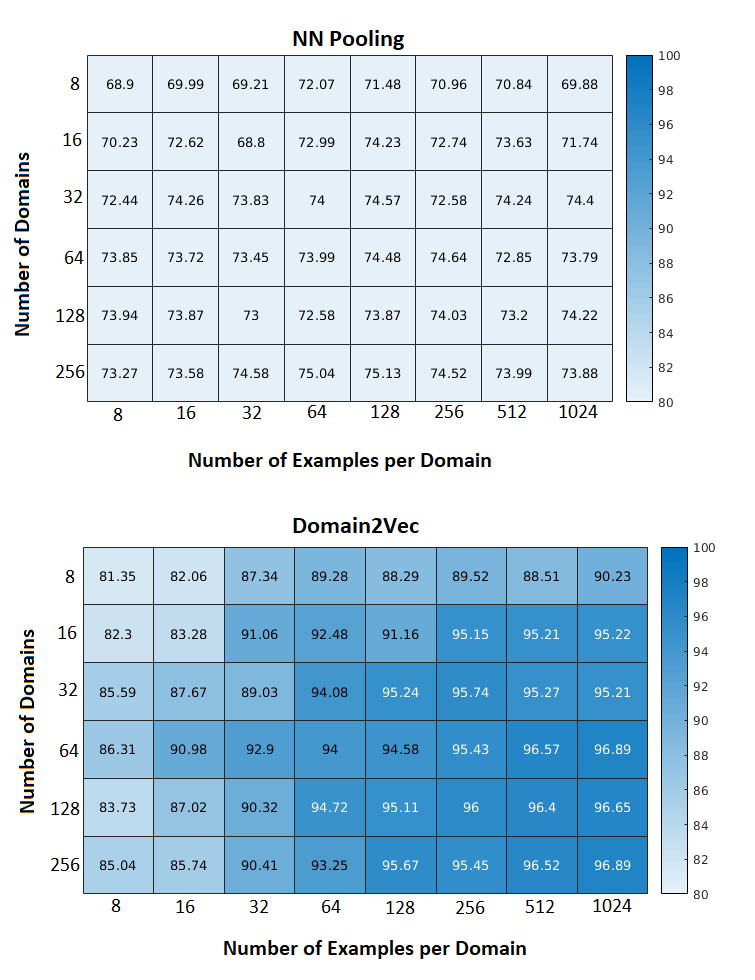}
     \caption{Heatmap comparing percentage accuracy on synthetic data (binary classification)}
     \label{fig:heatmap_synthetic}
\end{figure}

\begin{figure}[htb!]
 \centering
\begin{minipage}{0.25\linewidth}
  \centering
    \includegraphics[width=0.9\textwidth]{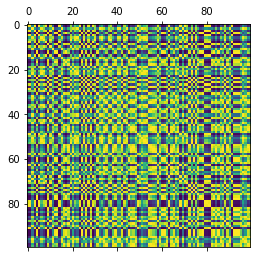}
     \caption{Known Task Similarity }
     \label{fig:known}
\end{minipage}
\hspace{0.3cm}
 \begin{minipage}{0.25\linewidth}
  \centering
    \includegraphics[width=0.9\textwidth]{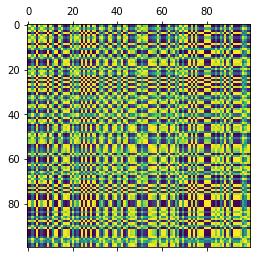}
     \caption{Estimated Task Similarity}
     \label{fig:synth}
\end{minipage}
\hspace{0.3cm}
 \begin{minipage}{0.25\linewidth}
  \centering
    \includegraphics[width=0.9\textwidth]{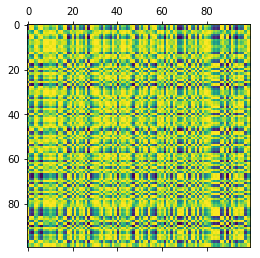}
     \caption{Random Task Similarity}
     \label{fig:random}
\end{minipage}
\end{figure}

We implement the architecture in TensorFlow and test the results on a 2-dimensional synthetic dataset and one real dataset. For synthetic dataset, we sample data points uniformly from a 2-D unit box \( [-1,1] \times [0,1] \). Data points are labeled according to their first dimension, i.e. \(1\) if it's non negative and \( 0 \) if it's negative. We then randomly sample \( \theta \) from \( [ 0, \pi] \) and  rotate all these data points by  \( \theta \) to get dataset of one domain. Note that when 
\( \theta \) of two different domains are close to each other then their data distribution and their decision functions are also very close to each other. 
In Fig. \ref{fig:heatmap_synthetic}, we show the result on synthetic dataset by varying number of training domains from 8 to 256 and number of examples per domain from 8 to 1024. Number of test domain are fixed to 44 and number of examples per test domain are fixed to 1024. We compare the results with a pooling Neural Network (pooling NN or 1 HNN), which combines data from all domains and trains one single neural network. Domain2Vec improves the accuracy from 73\% to 96\% compared to Pooling NN. Also, accuracy of Domain2Vec model increases as it gets more number of training example per domain and more number of training domains. 

We also show similarity between embeddings learnt by Domain2Vec for different domains and its comparison to true similarity (based on \( \theta \) ) and random similarity. Let \( d_p \) and \( d_q \) denote the learnt  embeddings for domain \( p \) and domain \( q \) respectively. Then, estimated similarity between domain \( p \) and domain \( q \) is \( \exp(-\frac{\| d_p - d_q \|^2}{\sigma^2}) \). Let \( \theta_p \) and  \( \theta_q \) denote the angles with which domain \( p \) and domain \( q \)  were created respectively. In this case, known similarity is  \( \exp(-\frac{| \theta_p - \theta_q |^2}{\sigma^2}) \). From the Fig. \ref{fig:synth}, it can be seen that similarity calculated using embeddings is very close to known similarity.

We provide the results on a popular dataset in domain generalization community called VLCS which consists of 4 different datasets: VOC2007  (V),  LabelMe  (L), Caltech-101  (C),  and  SUN09  (S).
We evaluate the classification accuracy of our model on the VLCS dataset by following the \textit{leave-one-domain-out} methodology discussed in \cite{ghifary2015domain}. Specifically, we train our model on three out of four datasets (source datasets) and test it on the fourth dataset (target dataset).  We follow the same experimental procedure as in \cite{gan2016learning} \footnote{Code is available here: https://github.com/aniketde/Domain2Vec}. From Table \ref{tb:results}, we can see that Domain2Vec improves the accuracy over all baselines. 
\begin{table}[htb!]
\centering
\begin{tabular}{ | p{10mm} | p{8mm} | p{11mm} | p{11mm} | p{16mm}| p{15mm} | p{15mm} | p{13mm} | p{14mm} | }
\hline
Source & Target & L-SVM & 1 HNN & Undo-Bias \cite{khosla2012undoing} & MTAE \cite{gan2016learning}& D-MTAE \cite{gan2016learning} & MLP \cite{li2017deeper} & D2V   \\ \hline
L,C,S & V &  58.86 & 59.10 & 54.29 & 61.10 & 63.90 &   65.58 & \textbf{65.94} \\ \hline
V,C,S & L &  52.49 & 58.20 & 58.09 & 59.24 & 60.13 & 58.74 & \textbf{60.28} \\ \hline
V,L,S & C &  77.67 & 86.67 & 87.50 & 90.71 & 89.05 & 92.43 & \textbf{92.50}  \\ \hline
V,L,C & S&  49.09 & 57.86 & 54.21 & 60.20 & 61.33 & 61.85 & \textbf{62.02}  \\ \hline
Average &  &  59.63 & 65.46 & 63.52 & 67.81 & 68.60 &69.65 &\textbf{70.82}      \\ \hline
\end{tabular}
\caption{Percentage Accuracy on VLCS}
\label{tb:results}
\end{table}

\section{Conclusion}
We address the problem of domain generalization and propose a neural network architecture to learn domain-specific embedding and then use this embedding along with data point specific features to label it. We show the effectiveness of the architecture by learning domain-specific embeddings and then accurately estimating domain to domain similarity. We outperform other state of the art algorithms on a standard domain generalization dataset. Extensive empirical studies are needed to make the domain embedding model more robust. Domain2Vec model can be easily extended to multi-task learning with homogeneous tasks \cite{ruder2017learning}. In future, we plan to devise a architecture similar to Domain2Vec that can learn domain-specific embeddings for other data types like text and speech.  

\bibliographystyle{abbrv}
\bibliography{references}

\end{document}